\documentclass[letterpaper, 10 pt, conference]{ieeeconf}  

\usepackage{graphicx}
\usepackage{blindtext}
\usepackage{hyperref}
\usepackage{gensymb}
\usepackage{amsmath}
\usepackage{amssymb}
\usepackage[utf8]{inputenc}
\usepackage{algorithm}
\usepackage{algpseudocode}
\usepackage{array}
\usepackage{mathrsfs}
\graphicspath{ {./images/} }
\DeclareMathOperator*{\argmin}{arg\,min}

\IEEEoverridecommandlockouts                              

\title{\LARGE \bf
SwarmCVT: Centroidal Voronoi Tessellation-Based Path Planning for Very-Large-Scale Robotics
}

\author{James Gao$^{1}$, Jacob Lee$^{1}$, Yuting Zhou$^{2}$, Yunze Hu,$^{3}$ Chang Liu$^{3}$, Pingping Zhu$^{1*}$
\thanks{This research was supported by the Defense Advanced Research Projects Agency (DARPA) -Grant \#000825 and the NASA Established Program to Stimulate Competitive Research (EPSCoR)-Grant \#80NSSC22M0027}
\thanks{$^{1}$James Gao, Jacob Lee, and Pingping Zhu are with the Department of Computer Sciences and Electrical Engineering (CSEE), Marshall University, Huntington, WV 25755, USA (gao32@marshall.edu; lee395@marshall.edu; zhup@marshall.edu).}
\thanks{$^{2}$Yuting Zhou is with the College of Health Professions, Marshall University, Huntington, WV 25755, USA (zhou54@marshall.edu)}
\thanks{$^{3}$Yunze Hu and Chang Liu are with the Department of Advanced Manufacturing and Robotics, College of Engineering, Peking University, Beijing 100871, China (hu\_yun\_ze@stu.pku.edu.cn; changliucoe@pku.edu.cn).}
\thanks{All correspondences should be sent to Pingping Zhu.}}

\begin{document}

\maketitle
\thispagestyle{empty}
\pagestyle{empty}

\begin{abstract}
Swarm robotics, or very large-scale robotics (VLSR), has many meaningful applications for complicated tasks. However, the complexity of motion control and energy costs stack up quickly as the number of robots increases. In addressing this problem, our previous studies have formulated various methods employing macroscopic and microscopic approaches. These methods enable microscopic robots to adhere to a reference Gaussian mixture model (GMM) distribution observed at the macroscopic scale. As a result, optimizing the macroscopic level will result in an optimal overall result. However, all these methods require systematic and global generation of Gaussian components (GCs) within obstacle-free areas to construct the GMM trajectories. This work utilizes centroidal Voronoi tessellation to generate GCs methodically. Consequently, it demonstrates performance improvement while also ensuring consistency and reliability.
\end{abstract}

\section{INTRODUCTION}
Path planning problems for very large-scale robotics (VLSR) systems are crucial research areas with broad applications from autonomous drone swarms to extensive fleets of self-driving vehicles and industrial robots \cite{lin2022review}. As the number of robots in a system increases, the challenge of coordinating efficient, collision-free paths grows significantly, requiring robust algorithms capable of handling complicated environments and real-time constraints. Effective path planning is essential for optimizing resource utilization, reducing energy consumption, and improving system performance. These advancements are particularly relevant in disaster relief, environmental monitoring, and intelligent transportation systems, where large-scale coordination is vital for mission success and operational efficiency.

The VLSR system control method is typically categorized as centralized \cite{yu2018effective, zhu2021adaptive} or decentralized \cite{regev2018decentralized, zhu2019scalable}. In the centralized approach, all robots are controlled by a master entity, whereas in the decentralized approach, individual robots or small clusters make decisions. The choice of methodology depends on resource availability and field applications.

This paper primarily focuses on applying the centralized method in a known obstacle workspace to evaluate its performance. The central concept involves dividing the path planning problem into macroscopic and microscopic scales. At the macroscopic level, the objective is to determine an optimal transport trajectory using a Gaussian mixture model (GMM). This trajectory can then serve as a reference for robots at the microscopic level to follow. The effectiveness of this approach heavily depends on the macroscopic path planning. Our previous works, both adaptive distributed optimal control (ADOC) \cite{zhu2021adaptive} and probabilistic roadmap motion planning for large-scale swarm robotic systems (swarmPRM) \cite{hu2024swarmprm}, have utilized this strategy. Extensive comparative tests with other existing methods for VLSR path planning have shown exceptional results. Therefore, this work aims to enhance our previous achievements further.

ADOC and swarmPRM methods require a systematic approach for generating Gaussian components to construct Gaussian Mixture Models (GMMs). In ADOC, the Gaussian components (GCs) are generated at predefined locations with identical covariance matrices, including those positioned within obstacles. To prevent collisions between robots and obstacles, ADOC applies a penalty term, which indicates the probability that robots are deployed within the obstacle areas. However, this approach leads to inefficiencies, increasing time and energy consumption, as these collocation GCs do not adapt dynamically to environmental obstacles. In contrast, swarmPRM generates GCs exclusively within obstacle-free regions, but the inherent randomization in this process introduces significant uncertainty in the resulting path planning.

To efficiently cover obstacle-free workspaces while minimizing overlap with obstacle regions, generating Gaussian components (GCs) systematically and globally is essential. To address this challenge, we propose a novel Gaussian distribution-based centroidal Voronoi tessellation (GCVT) approach, inspired by the traditional CVT method, to create a set of GCs that evenly cover the obstacle-free areas. Based on GCVT, we further develop a robotic swarm path planning algorithm named SwarmCVT. Comparative performance evaluations between SwarmCVT, ADOC, and SwarmPRM show significant numerical improvements with SwarmCVT, positioning it as a robust solution for tackling the VLSR path planning problem. The main contributions of this work are summarized as follows:

1) Propose the concept and formal definition of GCVT and develop two approximation methods to solve the GCVT problem ( see Algorithm-\ref{alg:cvt_gen}). 

2) Develop the SwarmCVT approach to the path planning problems for the VLSR systems in environments with known deployed obstacles (see Algorithm-\ref{alg:SwarmCVT}).  

3) Evaluate the proposed SwarmCVT approach and compare it with SwarmCVT and ADOC approaches, demonstrating the significant numerical improvements of SwarmCVT (see Section \ref{sec:Simulation_Results}).

\section{Problem Formulation}
Consider the path planning problem for a VLSR system consisting of $N$ homogeneous cooperative robots deployed in a large obstacle-populated region of interest (ROI) denoted by $\mathcal{W} \subset \mathbb{R}^2$. To simplify the problem, this paper assumes that the layout of all $M$ obstacles in the ROI is known a prior and stay static throughout, where the area occupied by the $m$th obstacle is represented by the set in the ROI, such that  $\mathcal{B}_m \subset \mathcal{W}$ for $m = 1, \ldots, M$. Then, we can denote the whole obstacle occupied areas by $\mathcal{B} = \bigcup\limits_{m=1}^{M}\mathcal{B}_{m} \subset \mathcal{W}$, and denote the non-occupied areas by $\mathcal{X} = \{ \mathbf{x} \in \mathcal{W} : \mathbf{x} \notin \mathcal{B}  \}$, which is the relative complement of $\mathcal{B}$ with respect to $\mathcal{W}$.      

We assume that the dynamics of the homogeneous robots can all be modeled by a general stochastic differential equation (SDE), such that
\begin{align}
	\dot{\mathbf{x}}_i(t) &= \mathbf{f}\left[\mathbf{x}_i(t), \mathbf{u}_i(t), \mathbf{n}_i(t)\right] \label{equ:robot_control_fn}\\
	\mathbf{x}_i(t_0) &=\mathbf{x}_{i_{0}}, i=1,...,N
\end{align}  
where $\mathbf{x}_i (t)$, $\mathbf{u}_i(t)$, and $\mathbf{n}_i(t)$ denotes the $i$th robot's state, the control input, and the system model noise at time $t$, respectively.  In addition, $\mathbf{x}_{i_0}$ denotes the $i$th robot's initial state at time $t_0$. To simplify the problem, this paper considers that the robot's state is only the position of the robot at time $t$, such that $\mathbf{x}_i(t), \mathbf{x}_{i0} \in \mathcal{W}$ for $i=1,\ldots, N$. Also, the robot's state or position is assumed to be fully observable and known with negligible errors. 

\subsection{General Form of Objective Function}
Similar to \cite{zhu2021adaptive}, let the macroscopic behavior of the VLSR systems be represented by the time-varying probability density function (PDF) of robots denoted by $\wp(\mathbf{x},t) \in \mathcal{P}(\mathcal{W})$, where $\mathcal{P}(\mathcal{W})$ is the space of PDFs defined on $\mathcal{W}$. Thus, the performance of the path planning problem for the VLSR system over a time interval $\left[t_0,t_f\right]$ can be evaluated by an integral objective function 
\begin{equation} 
    J\left[\wp(\mathbf{x},t)\right] \triangleq \phi\left[\wp(t_f), \wp_f\right]+\int_{t_0}^{t_f} \mathcal{L}\left[\wp(\mathbf{x},t)\right] dt 
    \label{eq:cont_time_costFunc}
\end{equation}
which represents the macroscopic cost required for the robots to move from a given initial distribution $\wp_0 \in  \mathcal{P}(\mathcal{W})$ at the time $t_0$ to a desired distribution $\wp_f \in \mathcal{P}(\mathcal{W})$ at the final time $t_f$. Here, the functionals $\phi\left[\wp(t_f),\wp_f\right]$ and $\mathcal{L}\left[\wp(\mathbf{x},t)\right]$ denote the terminal cost and intermediate step cost, or ``Lagrangian", respectively. Notably, this cost function is very similar to the one provided in \cite{zhu2021adaptive}, except that the information on obstacle layout is implicit since this paper considers a fixed obstacle layout for simplicity. 

Next, considering the algorithm implementation, we discretize the cost function with respect to time.  Let $\triangle t$ denote a small time interval such that $\triangle t \ll (t_f - t_0)$, and divide the whole time range into $T = \lceil(t_f - t_0) / \triangle t \rceil$ time steps, indexed by $t_k = t_0 + k \triangle t$, where $k = 1, \ldots, T$. Therefore, the objective functional in (\ref{eq:cont_time_costFunc}) can be approximated by
\begin{equation}
	J(\boldsymbol{\wp}_{0:T}) \triangleq \phi\left(\wp_{T}, \wp_f \right) + \sum_{k=0}^{T - 1}  \mathcal{L}(\wp_k, \wp_{k+1}) 
	\label{eq:discrete_time_costFunc}
\end{equation}d
where $\wp_k = \wp(\mathbf{x},k \triangle t)$ indicates the robots' PDF at the $k$th time step, and $\boldsymbol{\wp}_{0:T} = [\wp_0 \ldots \wp_{T}]$ denotes the sequence of the robot's PDF trajectory. Here, the term $\mathcal{L}(\wp_k, \wp_{k+1})$ is the discrete-time Lagrangian term, which is analog to the integration of $\mathcal{L}\left[\wp(\mathbf{x},t)\right]$ during one time step $\triangle t$, but not a simple and direct integration. Similar to \cite{yang2023risk, yang2024rover, hu2024swarmprm}, we formulate the terminal term and the Lagrangian term both based on the Wasserstein metric or Wasserstein distance. 

\subsection{Brief Introduction of Wasserstein Metric}       
The Wasserstein metric is a critical concept in optimal mass theory (OMT) \cite{chen2018OMT, zhu2021adaptive}, which can be applied to measure the distance between two distributions.  Specifically, for two Gaussian distributions associated with the PDFs, $g_1 \sim \mathcal{N}(\boldsymbol{\mu}_1, \boldsymbol{\Sigma}_1)$ and $g_2 \sim \mathcal{N}(\boldsymbol{\mu}_2, \boldsymbol{\Sigma}_2)$, the following closed form can calculate the Wasserstein metric,
\begin{align}
	W_2(g_1, g_2) &= \large\left\{ ||\boldsymbol{\mu}_1 - \boldsymbol{\mu}_2||^2 \right. \nonumber\\
	&+ \left. \text{tr} \left[\boldsymbol{\Sigma}_1 + \boldsymbol{\Sigma}_2 - 2 \left(\boldsymbol{\Sigma}_1^{\frac{1}{2}} \boldsymbol{\Sigma}_2 \boldsymbol{\Sigma}_1^{\frac{1}{2}}\right)^{1/2} \right] \right\}^{1/2}
\end{align}
In addition, the geodesic path or displacement interpolation from $g_1$ to $g_2$ is also a Gaussian distribution denoted by $g_{1,2}(\tau) \sim \mathcal{N}(\boldsymbol{\mu}(\tau), \boldsymbol{\Sigma}(\tau))$ for $0 \leq \tau \leq 1$, such that
\begin{align}
	\boldsymbol{\mu}(\tau) &= (1 - \tau)\boldsymbol{\mu}_1+ \tau \boldsymbol{\mu}_2 \label{eq:mu_interpolation} \\
	\boldsymbol{\Sigma}(\tau) &= \boldsymbol{\Sigma}_1^{-\frac{1}{2}}\left[(1 - \tau)\boldsymbol{\Sigma}_1 + \tau (\boldsymbol{\Sigma}_1^{\frac{1}{2}}\boldsymbol{\Sigma}_2 \boldsymbol{\Sigma}_1^{\frac{1}{2}})^\frac{1}{2}\right]^2 \boldsymbol{\Sigma}_1^{-\frac{1}{2}} \label{eq:Sigma_interpolation}
\end{align}
Therefore, a space of Gaussian distributions is equipped with the $W_2$ metric, denoted by $\mathcal{G}$. 

Moreover, although there is no efficient calculation for the Wasserstein metric for general distributions, a new Wasserstein-like metric was proposed to efficiently approximate the Wasserstein metric between two GMMs referred to as the Wasserstein-GMM (WG) metric, which is defined by
\begin{align}
	d(\wp_1, \wp_2) &\triangleq \left\{ \underset{\pi \in \Pi(\boldsymbol{\omega}_1, \boldsymbol{\omega}_2)}{\min}  \sum_{\imath=1}^{N_1}\sum_{\jmath=1}^{N_2}\left[W_2(g_1^{\imath},g_2^{\jmath})\right]^2 \pi_{1,2}(\imath,\jmath) \right\}^{\frac{1}{2}} \nonumber\\
	&= \tilde{d}(\wp_1, \wp_2, \pi^*_{1,2}) 
	\label{eq:WG_metric}
\end{align}
where 
\begin{equation}
	\tilde{d}(\wp_1, \wp_2, \pi_{1,2}) \triangleq \left\{\sum_{\imath=1}^{N_1}\sum_{\jmath=1}^{N_2}\left[W_2(g_1^{\imath},g_2^{\jmath})\right]^2 \pi_{1,2}(\imath,\jmath) \right\}^{1/2} 
	\label{eq:tilde_WG_metric}
\end{equation}   
Here, $\wp_1 = \sum_{\imath = 1}^{N_1} \omega_1^{\imath} g_1^{\imath}$ and  $\wp_2 = \sum_{\jmath = 1}^{N_1} \omega_2^{\jmath} g_2^{\jmath}$ are GMMs specified by the Gaussian components, $\{g_1^{\imath}\}_{\imath = 1}^{N_1}$ and $ \{g_2^{\jmath} \}_{\jmath = 1}^{N_2}$, and the corresponding weights, $\boldsymbol{\omega}_1 = [\omega_1^{1},\ldots, \omega_1^{\imath}, \ldots, \omega_1^{N_1}]$ and $\boldsymbol{\omega}_2 = [\omega_2^{1},\ldots, \omega_2^{\jmath}, \ldots, \omega_1^{N_2}]$, respectively. In addition, $\Pi(\boldsymbol{\omega}_1, \boldsymbol{\omega}_2)$ denote the space of joint discrete distributions with the maginals $\boldsymbol{\omega}_1$ and $\boldsymbol{\omega}_2$, and $\pi^*_{1,2}(\imath, \jmath)$ denotes the minimizer. Therefore, the space of GMMs equipped with the WG metric is a metric space referred to as Wasserstein-GMM space and denoted by $\mathcal{GM}$. Furthermore, a geodesic on $\mathcal{GM}$ connecting $\wp_1 \in \mathcal{GM}$ and $\wp_2 \in \mathcal{GM}$ is given by \cite{chen2018OMT}
\begin{equation}
	\wp_{1,2}(\tau) = \sum_{\imath=1}^{N_1}\sum_{\jmath=1}^{N_2} \pi^*_{1,2}(\imath,\jmath) g_{1,2}^{\imath, \jmath}(\tau) \text{ for } 0 \leq \tau \leq 1
\end{equation} 
where $g_{1,2}^{\imath, \jmath}(t) \in \mathcal{G}$ is the displacement interpolation between $g_1^{\imath}$ and $g_2^{\jmath}$ using (\ref{eq:mu_interpolation}) and (\ref{eq:Sigma_interpolation}). In addition, we can have
\begin{equation}
	d\left( \wp_{1,2}(\tau_1), \wp_{1,2}(\tau_2)  \right) = (\tau_2 - \tau_1) d( \wp_1, \wp_2 )
\end{equation}
It is noteworthy that Gaussian distributions can be treated as special GMMs with one component, such that $g_1, g_2 \in \mathcal{G} \subset \mathcal{GM}$, and $d(g_1, g_2) = W_2(g_1, g_2)$.

\subsection{Problem Formulation In Wasserstein-GMM Space}
Without loss of generality, we can approximate the time-varying robots' PDFs and the desired PDF are all GMMs, such that $\wp_k, \wp_f \in \mathcal{G}$, where $\wp_k = \sum_{\imath = 1}^{N_k} \omega_k^{\imath} g_k^{\imath}$ and $\wp_f = \sum_{\jmath = 1}^{N_f} \omega_k^{\jmath} g_k^{\jmath}$, for $k = 0, 1, \ldots, T$. Because the W2 and WG distances are both $l2$ norms, then it is straightforward to define the terminal and Lagrangian terms in (\ref{eq:discrete_time_costFunc}) as the square of the corresponding WG distances, such that
\begin{align}
	\phi\left(\wp_{T}, \wp_f \right) &= d^2(\wp_{T}, \wp_f)\\
	\mathcal{L}(\wp_k, \wp_{k+1}) &= d^2(\wp_k, \wp_{k+1})
\end{align}     
where $\phi\left(\wp_{T}, \wp_f \right)$ and $\mathcal{L}(\wp_k, \wp_{k+1})$ are both the linear combinations of the squared W2 distances of their Gaussian components. 

Then, the objective function in (\ref{eq:discrete_time_costFunc}) can be rewritten by
\begin{equation}
	J(\boldsymbol{\wp}_{0:T}) \triangleq d^2\left(\wp_{T}, \wp_f \right) + \sum_{k=0}^{T - 1}  d^2(\wp_k, \wp_{k+1}) 
	\label{eq:discrete_time_costFunc_WG}
\end{equation}
Because the step interval $\triangle T$ is a fixed, the term $d^2(\wp_k, \wp_{k+1})$ is proportional to the square of the distribution velocity, and the cost function $J(\boldsymbol{\wp}_{0:T})$ in (\ref{eq:discrete_time_costFunc_WG}) represent the energy cost from $\wp_0$ to  $\wp_f$. Therefore, the objective function in (\ref{eq:cont_time_costFunc})  is approximated and reformulated as a special shortest path-planning problem in the Wasserstein-GMM space, where the objective is to reduce the energy cost rather than the length of trajectory.   

This objective function in (\ref{eq:discrete_time_costFunc_WG}) is very similar to the one provided in \cite{zhu2021adaptive} except that a penalty term, $p(\wp_{k+1})$, was added to the Lagrangian term in \cite{zhu2021adaptive}, such that  $\mathcal{L}_{ADOC}(\wp_k, \wp_{k+1}) = d^2(\wp_k, \wp_{k+1}) + p_{\mathcal{B}}(\wp_{k+1})$. The penalty term is defined by
\begin{equation}
	p_{\mathcal{B}}(\wp) = \int_{\mathbf{x} \in \mathcal{B} } \wp(\mathbf{x}) d\mathbf{x}
\end{equation}
which is referred to as the distribution avoidance penalty (DAP) and indicates the probability that robots are deployed within the areas occupied by obstacles. 

In order to introduce the absented penalty term $p_{\mathcal{B}}$ to the objective function, we find out 
the optimal trajectory of the robot distributions for the VLSR system by solving the following optimization problem with constraints such that   
\begin{align}
	\boldsymbol{\wp}_{0:T_f}^* &= \underset{{\wp}_{0:T_f}}{\min}J(\boldsymbol{\wp}_{0:T_f}) \nonumber\\
	\text{s.t.   } &  p_{\mathcal{B}} (\wp_k) < \eta_{\mathcal{B}} \text{ for } k = 1,\ldots, T \label{eq:optimization_constraint}
\end{align}  
where $0 < \eta_{\mathcal{B}} \ll 1$ is a user-defined threshold term indicating the maximum tolerable DAP.

\section{Path Planning In Wasserstein-GMM Space}
\label{sec:Path_Planning}
The constrained optimization provided in (\ref{eq:optimization_constraint}) describes the robotic swarm path planning problem in the Wasserstein-GMM space well in aspects of conception and theory. However, the optimal solution is not easy to solve. Because, first, too many variables are involved, including the parameters specifying these GMMs, $\wp_k$, for $k=1,\ldots, T$; second, the calculation of $d(\wp_k, \wp_{k+1})$ involves the minimization operator. This section provides a sub-optimal solution by optimizing an upper bound of the objective function in the Wasserstein-GMM space. 

\subsection{Upper Bound of Objective Function}
The objective function $J(\boldsymbol{\wp}_{0:T})$ in (\ref{eq:discrete_time_costFunc_WG}) is complex because the WG metric, $d(\cdot, \cdot)$, is defined based on the minimum operator in (\ref{eq:WG_metric}). To remove the minimum operator in the objective function, we can replace $d(\cdot, \cdot)$ with $\tilde{d}(\cdot, \cdot,\cdot)$, which is defined in (\ref{eq:tilde_WG_metric}). Then, we have a new objective function denoted by $\tilde{J}$, such that
\begin{align}
	&\tilde{J}(\boldsymbol{\wp}_{0:T}, \boldsymbol{\pi}_{0:T}, \pi_{T,f}) \nonumber\\
	\triangleq & \tilde{d}^2\left(\wp_{T}, \wp_f, \pi_{T,f} \right) 
	+ \sum_{k=0}^{T - 1}  \tilde{d}^2(\wp_k, \wp_{k+1}, \pi_{k, k+1})
\end{align} 
where $\boldsymbol{\pi}_{0:T} = [\pi_{0,1} \ldots \wp_{T-1,T}]$ denotes the sequence of the joint distributions, $\pi_{k, k+1}$ for $k = 0,\ldots,T-1$. It is obvious given $\boldsymbol{\wp}_{0:T}$, $\tilde{J}(\boldsymbol{\wp}_{0:T}, \boldsymbol{\pi}_{0:T}, \pi_{T,f})$ is an upper bound of the objective function in (\ref{eq:discrete_time_costFunc_WG}), such that   
\begin{align}
	J(\boldsymbol{\wp}_{0:T}) = \tilde{J}(\boldsymbol{\wp}_{0:T}, \boldsymbol{\pi}^*_{0:T}, \pi^*_{T,f}) \leq \tilde{J}(\boldsymbol{\wp}_{0:T}, \boldsymbol{\pi}_{0:T}, \pi_{T,f}) \label{eq:Upper_bound}
\end{align}

Although the minimum operator is removed, more parameters are introduced into $\tilde{J}$, which results in the optimization problem is still complicated to solve. Similar to  \cite{zhu2021adaptive, hu2024swarmprm}, to simplify the optimization problem, we make the following assumptions on  the number of Gaussian components and these joint distributions for the robots' PDFs,  $\boldsymbol{\wp}_{0:T}$, 
\begin{align}
	\wp_T &= \wp_f  \label{eq:ass-1}\\
	N_k &= N_0 \times N_f   \text{ for } k = 1, \ldots, T-1  \label{eq:ass-2} \\
	\pi_{k,k+1}(\imath, \jmath) &= \begin{cases}
		\omega_k^{\imath}, &\text{if } \imath = \jmath, \\
		0, &\text{otherwise }
	\end{cases}  \text{for } 1 \leq k \leq T-2 \label{eq:ass-3}\\
	\pi_{0,1}(\imath, \jmath) &= 0 \text{ if } \imath \neq \lceil \jmath / N_0 \rceil \label{eq:ass-4}\\
	\pi_{T-1, T}(\imath, \jmath) &= 0 \text{ if } \jmath \neq \text{mod}(\imath, N_T) \label{eq:ass-5}
\end{align}
where ``mod" indicates the modulo operator. With the above assumptions (\ref{eq:ass-1}) - (\ref{eq:ass-5}), there exists a unique trajectory of Gaussian components (GCs) from $g_0^{\imath}$ to $g_T^{\jmath}$ for $1 \leq \imath \leq N_0$ and $1 \leq \jmath \leq N_T$ which is denoted by $\mathcal{T}_{0,T}^{\imath, \jmath} = [g_0^{\imath}, \ldots, g_k^{\imath, \jmath}, \ldots,  g_T^{\jmath}]$. Then, the distribution trajectory from $\wp_0$ to $\wp_f$ can be divided into $N_0 * N_f$  trajectories from the GCs $\{g_0^{\imath} \}_{\imath = 1}^{N_0}$ to the GCs $\{g_f^{\jmath} \}_{\jmath = 1}^{N_f}$ in total. 

According to (\ref{eq:Upper_bound}), the cost function generated under the assumptions (\ref{eq:ass-1}) - (\ref{eq:ass-5}) is still an upper bound of $J(\boldsymbol{\wp}_{0:T}) $. Therefore, we can redefine the upper bound of the objective function by
\begin{align}
	\tilde{J}(\boldsymbol{\wp}_{0:T}) \triangleq \sum_{\imath = 1}^{N_0}\sum_{\jmath = 1}^{N_T} \omega_{\imath, \jmath} \mathcal{L} (g_0^{\imath}, g_T^{\jmath})
	\label{eq:upper_bound_objective_function}
\end{align} 
where $\mathcal{L} (g_0^{\imath}, g_T^{\jmath})$ is the cost of the trajectory $\mathcal{T}_{0:T}^{\imath, \jmath}$, such that
\begin{align}
	\mathcal{L} (g_0^{\imath}, g_T^{\jmath}) &\triangleq \left[W_2(g_0^{\imath}, g_1^{\imath, \jmath})\right]^2 + \sum_{k=1}^{T-2} \left[W_2(g_k^{\imath, \jmath}, g_{k+1}^{\imath, \jmath})\right]^2 \nonumber\\
	&+ \left[W_2(g_{T-1}^{\imath, \jmath}, g_T^{\jmath})\right]^2
\end{align}
and $\omega_{\imath, \jmath}$ indicates the weights for the corresponding trajectories. Because of the assumptions (\ref{eq:ass-1}) and (\ref{eq:ass-3}), we can express the GMMs $\wp_k$ for $1 \leq k \leq T-1$ by
\begin{equation}
	\wp_k = \sum_{\imath = 1}^{N_0}\sum_{\jmath = 1}^{N_0} \omega_{\imath, \jmath} g_k^{\imath, \jmath} = \sum_{\imath = 1}^{N_0}\sum_{\jmath = 1}^{N_f} \omega_{\imath, \jmath} g_k^{\imath, \jmath}
	\label{eq:wp_k}
\end{equation}

\subsection{Constrained Optimization in Wasserstein-GMM Space}
At this point, we can approximate the optimal solution of the problem in (\ref{eq:discrete_time_costFunc_WG}) by minimizing the upper bound of the objective function defined in (\ref{eq:upper_bound_objective_function}). Because the terms $\omega_{\imath, \jmath}$ and $\mathcal{L}(g_0^{\imath}, g_T^{\jmath})$ are uncoupled, this optimization can be solved in two steps. Step-1 is to minimize the cost of the trajectory $\mathcal{T}_{0:T}^{\imath, \jmath}$ for every pair of GCs, $\{(g_0^{\imath}, g_f^{\jmath})\}_{\imath = 1, \jmath = 1}^{N_0, N_f}$, which is denoted by $\mathcal{L}_{0:T}^{*\imath, \jmath}$, such that 
\begin{equation}
	\mathcal{L}_{0:T}^{*\imath, \jmath} = \underset{\mathcal{T}_{0:T}^{*\imath, \jmath}}{\min} 	\mathcal{L} (g_0^{\imath}, g_T^{\jmath})
\end{equation}
This step can be implemented in a parallel fashion by using the shortest path planning algorithms in the space $\mathcal{G}(\mathcal{W})$. Step-2, then, is to optimize the weights $\{\omega_{\imath, \jmath}\}_{\imath = 1, \jmath = 1}^{N_0, N_f}$ given the optimal trajectory cost $\{\mathcal{L}_{0:T}^{*\imath, \jmath}\}_{\imath = 1, \jmath = 1}^{N_0, N_f}$ obtained in Step-1. 

Finally, since $\wp_k$ is represented as a linear combination of GCs in (\ref{eq:wp_k}), the path planning in Step-1 can be executed within a subset of $\mathcal{G}(\mathcal{W})$, ensuring that the resulting robots' PDFs, $\wp_k$, meet the constraints, $p_{\mathcal{B}} (\wp_k) < \eta_{\mathcal{B}}$, specified in (\ref{eq:discrete_time_costFunc_WG}). This subset, $\tilde{\mathcal{G}}(\mathcal{W}) \subset \mathcal{G}(\mathcal{W})$ is defined as follows,
\begin{equation} 
	\tilde{\mathcal{G}}(\mathcal{W}) \triangleq \left\{g \vert g \in \mathcal{G}(\mathcal{W}), p_{\mathcal{B}} (g) < \eta_{\mathcal{B}} \right\} 
\end{equation}

\section{Gaussian Distribution Based Centroidal Voronoi Tessellation}
In this section, our aim is to generate a set of GCs in $\tilde{\mathcal{G}}(\mathcal{W})$, referred to as collocation GCs, denoted as $G_C = \{g_c^\imath\}_{\imath = 1}^{K} \subset \tilde{\mathcal{G}}$, similar to those proposed in \cite{zhu2021adaptive}. These collocation GCs serve as nodes for constructing graphs to determine the shortest trajectories $\mathcal{T}_{0:T}^{\imath, \jmath}$ for $\imath = 1, \ldots, N_0$ and $\jmath = 1, \ldots, N_f$. With the workspace $\mathcal{W}$ and obstacle-free areas $\mathcal{X}$ fully defined, generating well-distributed collocation GCs within $\mathcal{X}$ is straightforward, ensuring that each component occupies a roughly uniform area. Since Centroidal Voronoi Tessellation (CVT) \cite{du2010advances} is a technique for creating regions of similar size, uniformly distributed across the workspace, we extend the traditional CVT into the Wasserstein-GMM space and develop a Gaussian distribution-based CVT to generate the required collocation GCs.
 
\subsection{Traditional Centroidal Voronoi Tessellation}
The traditional CVT is defined in Euclidean space, especially $\mathcal{W} \subset \mathcal{R}^2$. Given $K>0$ distinct points, $\{ \mathbf{p}_i\}_{i=1}^K \subset \mathcal{W}$, we can define $K$ Voronoi regions (VRs), $\mathcal{V}_i \subset \mathcal{W}$, for $i, j = 1,\ldots, K$, by
\begin{equation} 
    \mathcal{V}_i \triangleq \left\{\mathbf{x} \in \mathcal{W} \left\vert \Vert\mathbf{x}-\mathbf{p}_i\Vert \le \Vert \mathbf{x}-\mathbf{p}_j \Vert  \text{ for } i \neq j  \right. \right\} \label{equ:voronoi_region_def}
\end{equation}
where the point $\mathbf{p}_i \in \mathcal{W}$ is referred to as a generator of the corresponding VR, $\mathcal{V}_i$. It is obvious that $\mathcal{V}_i \cap \mathcal{V}_j = \emptyset$, and the set of VRs, $\{ \mathcal{V}_i \}_{i=1}^K$ is a tessellation of $\mathcal{W}$, referred as the Voronoi tessellation (VT) of $\mathcal{W}$ \cite{du2010advances}. 

Furthermore, given a density function $\rho(\mathbf{x} ) \geq 0$ defined on $\mathcal{W}$, the centroid of the VR $\mathcal{V}_i$, denoted by $\mathbf{c}_i$, can be defined by
\begin{equation}
	\mathbf{c}_i \triangleq \frac{ \int_{\mathcal{V}_i} \mathbf{x} \rho(\mathbf{x}) d\mathbf{x} }{\int_{\mathcal{V}_i} \rho(\mathbf{x}) d\mathbf{x}}
	\label{eq:defition_centroid}
\end{equation}
Then, a Voronoi tessellation of $\mathcal{W}$, specified by $\boldsymbol{\Theta}^*_{CVT} = \{ \left(\mathbf{p}_i^*, \mathcal{V}^*_i \right ) \}_{i=1}^K$, is said to be a centroidal Voronoi tessellation (CVT) if and only if these generators are also the centroids of the corresponding VRs, such that $\mathbf{p}_i^* = \mathbf{c}_i$ for $i = 1,\ldots, K$. 

The following  clustering objective function can specify the optimization property of the CVTs, 
\begin{equation}
	J_{CVT}(\boldsymbol{\Theta}_{CVT}) = \sum_{i=1}^K \int_{ \mathcal{V}_i} \rho(\mathbf{x}) \Vert \mathbf{x} - \mathbf{p}_i \Vert^2 d\mathbf{x}
	\label{eq:objective_func_CVT}
\end{equation}
where the objective function $J_{CVT}$ is minimized only if $\boldsymbol{\Theta}^*_{CVT}$ forms a CVT of $\mathcal{W}$. Based on the objective function in (\ref{eq:objective_func_CVT}), many CVT variants are developed \cite{lloyd1982least, du2003constrained, arthur2006k, lee2024adaptive} in several different applications, including area coverage and path planning. Among these CVT variants, the CVT in the workspace with obstacles is worth mentioning. An easy and feasible method is to modify the density function by embedding the obstacle information. Specifically, the density function is redefined by 
\begin{equation}
	\rho(\mathbf{x}) \triangleq \begin{cases}
		0 & \text{ if } \mathbf{x} \in \mathcal{B}\\
		\rho(\mathbf{x}) & \text{ otherwise }
	\end{cases}
\end{equation} 
where $\mathcal{B} \subset \mathcal{W}$ indicates the areas occupied by obstacles. Fig. \ref{fig:cvt_result_comp} (a) demonstrates an example of CVT in a workspace with obstacles. 

\subsection{Gaussian Distribution-based CVT}
To generate a set of $K$ collocation GCs in $\tilde{\mathcal{G}}(\mathcal{W})$ to cover the entire free-obstacle areas in the workspace, $\mathcal{X} \subset \mathcal{W}$, we develop a novel CVT-based approach, where every VT, $\mathcal{V}_i$, is dominated by one unique GC, $g(\mathbf{x} \vert \boldsymbol{\theta}_i)$, which is specified by the parameter $\boldsymbol{\theta}_i = (\boldsymbol{\mu}_i, \boldsymbol{\Sigma}_i)$ for $i = 1, \ldots, K$. 

Unlike the traditional VT, which is only specified by one parameter $\mathbf{p}_i$, our VTs depend on two parameters, $\boldsymbol{\mu}_i$ and $\boldsymbol{\Sigma}_i$, first, we can redefine our VT by
\begin{align}
    \mathcal{V}_i &\triangleq \left\{\mathbf{x} \in \mathcal{W} \left\vert g(\mathbf{x} \vert \boldsymbol{\theta}_i)\geq  g(\mathbf{x} \vert \boldsymbol{\theta}_j)  \text{ for } i \neq j  \right. \right\} \label{equ:voronoi_region_def_1} \\
     &\triangleq \left\{\mathbf{x} \in \mathcal{W} \left\vert d_{\boldsymbol{\Sigma}_i}(\mathbf{x}, \boldsymbol{\mu}_i) \leq d_{\boldsymbol{\Sigma}_j}(\mathbf{x}, \boldsymbol{\mu}_j)  \text{ for } i \neq j  \right. \right\} 
     \label{equ:voronoi_region_def_2}
\end{align}  
where the term $d_{\boldsymbol{\Sigma}}(\mathbf{x}, \boldsymbol{\mu})$ is defined by
\begin{align}
	d_{\boldsymbol{\Sigma}}(\mathbf{x}, \boldsymbol{\mu}) &\triangleq  -\ln[ g(\mathbf{x} \vert \boldsymbol{\mu}, \boldsymbol{\Sigma}) ] \nonumber\\
	&= \frac{1}{2}(\mathbf{x} - \boldsymbol{\mu})^T \boldsymbol{\Sigma}^{-1}(\mathbf{x} - \boldsymbol{\mu}) + \ln \big( 2\pi \vert \boldsymbol{\Sigma}\vert^{\frac{1}{2}} \big)   \nonumber\\
	&= \frac{1}{2}\Vert \mathbf{x} - \boldsymbol{\mu}  \Vert^2_{\boldsymbol{\Sigma}} + \ln \big( 2\pi \vert \boldsymbol{\Sigma}\vert^{\frac{1}{2}} \big) 
\end{align}
Here,  $\Vert \mathbf{x} - \boldsymbol{\mu}  \Vert = \big[(\mathbf{x} - \boldsymbol{\mu})^T \boldsymbol{\Sigma}^{-1}(\mathbf{x} - \boldsymbol{\mu})\big]^{1/2}$ denotes the Mahalanobis distance with respect to $\boldsymbol{\Sigma}_i$ \cite{richter2015mahalanobis}. 

Next, because $\mathcal{V}_i$ is dominated by the GC $g(\mathbf{x} \vert \boldsymbol{\theta}_i)$, it is straightforward to redefine the density function by
\begin{equation}
	\rho(\mathbf{x}) \triangleq \begin{cases}
		0 & \text{ if } \mathbf{x} \in \mathcal{B}\\
		g(\mathbf{x} \vert \boldsymbol{\theta}_i) & \text{ if } \mathbf{x} \in \mathcal{V}_i \cap \mathcal{X} 
	\end{cases}
	\label{eq:definition_desity_function}
\end{equation} 
By substituting $\rho(\mathbf{x})$ defined in (\ref{eq:definition_desity_function}) into (\ref{eq:defition_centroid}), we can have the Gaussian distribution-based centroid, $\mathbf{c}_i \in \mathcal{V}_i$. In addition, it is noteworthy that we can also have $\tilde{\mathcal{V}}_i = \mathcal{V}_i \cap \mathcal{X}$ and $\{ \tilde{\mathcal{V}}_i \}_{i=1}^K$ is a tessellation of $\mathcal{X}$. Due to the definition of the density function in (\ref{eq:definition_desity_function}), we do not distinct $\mathcal{V}_i$ and $\tilde{\mathcal{V}}_i$ for $i = 1,\ldots, K$ from hereon. 

Then, we can define the new CVT based on the Gaussian distributions, referred to as Gaussian distribution-based CVT (GCVT), where $\boldsymbol{\mu}^*_i = \mathbf{c}_i$ for all $i = 1,\ldots, K$. Analogous to the traditional CVT, given $\mathcal{W}$ or $\mathcal{X}$, the GCVT is specified by $\boldsymbol{\Theta}^*_{GCVT} = \{ (\boldsymbol{\theta}^*_i,  \mathcal{V}^*_i) \}_{i=1}^K$, and can be obtained by minimizing the following objective function,
\begin{align}
	J_{GCVT}(\boldsymbol{\Theta}_{GCVT}) &\triangleq  \sum_{i=1}^K J_{GCVT,i}(\boldsymbol{\theta}_i, \mathcal{V}_i)  \label{eq: objective_func_GCVT}\\
	J_{GCVT,i}(\boldsymbol{\theta}_i, \mathcal{V}_i) &\triangleq  \int_{ \mathcal{V}_i} -g(\mathbf{x} \vert \boldsymbol{\theta}_i) \ln[ g(\mathbf{x} \vert \boldsymbol{\theta}_i) ] d\mathbf{x}  \label{eq: objective_func_GCVT_i}
\end{align}
where the Euclidean distance in (\ref{eq:objective_func_CVT}) is replaced by the term $d_{\boldsymbol{\Sigma}_i}(\mathbf{x}, \boldsymbol{\mu}_i)$ in (\ref{equ:voronoi_region_def_2}). 

Furthermore, the objective function defined in (\ref{eq: objective_func_GCVT}) can be reexpressed by,  
\begin{align}
	J_{GCVT}(\boldsymbol{\Theta}_{GCVT}) = \frac{1}{2} J_{\boldsymbol{\Sigma}} + J_P 
\end{align}
where 
\begin{align}
	J_{\boldsymbol{\Sigma}}(\boldsymbol{\Theta}_{GCVT}) & \triangleq \sum_{i=1}^K \int_{ \mathcal{V}_i} g(\mathbf{x} \vert \boldsymbol{\theta}_i) \Vert \mathbf{x} - \boldsymbol{\mu}_i \Vert^2_{\boldsymbol{\Sigma}_i} d\mathbf{x} \label{eq:costFunc_J_Sigma}\\
	J_P(\boldsymbol{\Theta}_{GCVT}) & \triangleq \sum_{i=1}^K   \ln \big(2\pi\vert \boldsymbol{\Sigma}_i \vert^{\frac{1}{2}} \big)  \int_{ \mathcal{V}_i} g(\mathbf{x} \vert \boldsymbol{\theta}_i) d\mathbf{x} \label{eq: costFunc_J_P}
\end{align}
Here, the first term, $J_{\boldsymbol{\Sigma}}(\boldsymbol{\Theta}_{GCVT})$, can be treated as a Mahalanobis CVT \cite{richter2015mahalanobis} associated with a special density function defined in (\ref{eq:definition_desity_function}). The second term is a weighted sum of the masses of all VTs, where these weights depend on the corresponding covariance matrices. Therefore, minimizing the objective function $J_{GCVT}(\boldsymbol{\Theta}_{GCVT})$ without extra constraints results in the
trivial solution where $\boldsymbol{\Sigma}_i = \mathbf{0}$ for $i = 1, \ldots, K$ \cite{richter2015mahalanobis}.   


Finally, recalling the goal of developing the GCVT, we can find the $K$ collocation GCs by solving the following constrained optimization problem, 
\begin{align}
	\boldsymbol{\Theta}^*_{GCVT} &= \underset{\boldsymbol{\Theta}_{GCVT}}{\argmin} J_{GCVT}(\boldsymbol{\Theta}_{GCVT}) \label{eq:argmin_GCVT} \\
	\text{s.t.   } &  p_{\mathcal{B}} (g(\mathbf{x}|\boldsymbol{\theta}_i)) < \eta_{\mathcal{B}} \text{ for } i = 1,\ldots, K  \label{eq:constraint_GCVT_obstacle}\\
	& \max_{\mathbf{x} \in \mathcal{V}_i} g(\mathbf{x}|\boldsymbol{\theta}_i) < \rho_{max} \text{ for } i = 1,\ldots, K \label{eq: constraint_GCVT_max_density}
\end{align} 
where $\rho_{max} > 0$ is a user-defined parameter indicating the maximum spatial density of the robots in the whole workspace, which is determined according to the number of robots and the individual robot's physical size. This constraint in (\ref{eq: constraint_GCVT_max_density}) is applied to determine the optimization, avoiding the trivial solution.


\subsection{Heuristic Approach for Gaussian-based CVT generation}

The constrained optimization problem described in (\ref{eq:constraint_GCVT_obstacle}) - (\ref{eq: constraint_GCVT_max_density}) can theoretically be solved using iterative methods, such as Lloyd’s algorithm \cite{lloyd1982least}. However, this problem is highly complex and computationally intensive due to the intricate coupling of the GCVT parameters, $\boldsymbol{\Theta}_{GCVT} = { (\boldsymbol{\mu}_i, \boldsymbol{\Sigma}_i, \mathcal{V}i) }{i=1}^K$. To address this complexity and improve computational efficiency, we propose a heuristic two-fold approach as a practical workaround for deriving the GCVT solution.

First, we introduce an initial assumption regarding the covariance matrices, $\{\boldsymbol{\Sigma}_i\}_{i=1}^K$, for the optimization process, such that
\begin{align}
	\boldsymbol{\Sigma}_i^0 &= \boldsymbol{\Sigma}_0  \text{ for } i = 1, \ldots, K \label{ass-GCVT_1}\\
	\lambda_1, \lambda_2  &\gg 0 \label{ass-GCVT_2}
\end{align}
where the superscript `0' denotes the initial guess at the $(l=0)$th iteration, and $\lambda_1$ and $\lambda_2$ indicate the eigenvalues of $\boldsymbol{\Sigma}_0$. Under this assumption, the density function $\rho(\mathbf{x}) = g(\mathbf{x} \vert \boldsymbol{\theta}_i) = \rho_0$ for $\mathbf{x} \in \mathcal{V}_i$ is treated as a constant, as long as $\boldsymbol{\mu}_i \in \mathcal{V}_i$ for $i = 1, \dots, K$. Consequently, the cost function in (\ref{eq: costFunc_J_P}) simplifies to a constant, $J_P(\boldsymbol{\Theta}_{GCVT})  =   \ln \big(2\pi\vert \boldsymbol{\Sigma}_0 \vert^{\frac{1}{2}} \big) \rho_0 \int_{ \mathcal{W}}  d\mathbf{x}$, and the cost function in (\ref{eq:costFunc_J_Sigma}) regraded to the standard CVT cost function defined in (\ref{eq:objective_func_CVT}), where the density function is constant, $\rho(\mathbf{x}) = \rho_0$ for $\mathbf{x} \in \mathcal{V}$. Using this assumption, we can obtain the set $\{( \boldsymbol{\mu}_i^0, \mathcal{V}_i^0 ) \}_{i=1}^K$ by applying state-of-the-art CVT algorithms, where $\boldsymbol{\mu}_i^0 = \mathbf{p}_i^*$. In this paper, we employ the k-means++ algorithm \cite{arthur2006k} due to its efficient and careful seeding approach.


Second, we approximate the parameters $\{(\boldsymbol{\mu}_i, \mathcal{V}i)\}_{i=1}^K$ by using the previously obtained parameters $\{(\boldsymbol{\mu}_i^0, \mathcal{V}i^0)\}_{i=1}^K$, denoted as $\{(\hat{\boldsymbol{\mu}}_i, \hat{\mathcal{V}}i)\}_{i=1}^K$. With these approximations, we proceed to optimize the cost function with respect to the covariance matrices $\{\boldsymbol{\Sigma}_i\}_{i=1}^K$. Since the regions $\mathcal{V}_i$ are approximated and the objective function has a summation form as in (\ref{eq: objective_func_GCVT}), the optimization process can be efficiently executed in parallel.

To speed up and simplify this optimization process, furthermore, we have the following approximation. Given that the GC $g(\mathbf{x} \vert \boldsymbol{\mathcal{\theta}}_i)$ dominate the VT $\mathcal{V}_i$, the objective function in (\ref{eq: objective_func_GCVT_i}) can be approximated by
\begin{align}
	J_{GCVT,i}(\boldsymbol{\theta}_i, \mathcal{V}_i) &\approx  \int_{ \mathbb{R}^2} -g(\mathbf{x} \vert \boldsymbol{\theta}_i) \ln[ g(\mathbf{x} \vert \boldsymbol{\theta}_i) ] d\mathbf{x} \nonumber\\
	& = H[g(\mathbf{x} \vert \boldsymbol{\theta}_i)] \nonumber\\
	& = \ln(2\pi) + 1 + \frac{1}{2} \ln(\vert  \boldsymbol{\Sigma}_i \vert ) \label{eq:approx_objective_func_GCVT_i}
\end{align}
where $H[g(\mathbf{x} \vert \boldsymbol{\theta}_i)]$ indicates the information entropy of the PDF $g(\mathbf{x} \vert \boldsymbol{\theta}_i)$.  By replacing the objective function, $J_{GCVT,i}(\boldsymbol{\theta}_i, \mathcal{V}_i)$,  by this approximation in (\ref{eq:approx_objective_func_GCVT_i}), we can approximate $\boldsymbol{\Sigma}_i$ through merely minimizing $\vert \boldsymbol{\Sigma}_i\vert$. It is noteworthy that the approximation is valid for a relatively small covariance matrix and may not fully align with our initial guess in (\ref{ass-GCVT_2}). 

Moreover, considering that the constraint of the VT is removed from the approximated and simplified objective function, we apply the following distinct initial guess $\hat{\boldsymbol{\Sigma}}_i^0$ for the $i$th VT, $\mathcal{V}_i$,   
\begin{equation}
	\hat{\boldsymbol{\Sigma}}_i^0 = \kappa \int_{\mathcal{V}_i} (\mathbf{x} - \boldsymbol{\mu}_i)^T (\mathbf{x} - \boldsymbol{\mu}_i) dx 
\end{equation} 
where $\kappa \gg 0$ is a user-defined parameter to satisfy the initial guess assumption in (\ref{ass-GCVT_2}). Therefore, we have the optimization problem as follows.
\begin{align}
	\hat{\boldsymbol{\Sigma}}_i &= \underset{\boldsymbol{\Sigma}_i}{\argmin} \vert  \boldsymbol{\Sigma}_i \vert \label{eq:argmin_GCVT_det_Sigma} \\
	\text{s.t.   } &  p_{\mathcal{B}} (g(\mathbf{x} \vert \hat{\boldsymbol{\mu}}_i, \boldsymbol{\Sigma}_i) < \eta_{\mathcal{B}}   \\
	& \int_{\mathcal{V}_i} g(\mathbf{x} \vert \hat{\boldsymbol{\mu}}_i, \boldsymbol{\Sigma}_i) \geq \eta_{\mathcal{V}}   \label{eq: constraint_GCVT_max_mass}
\end{align} 
where the parameter $\eta_{\mathcal{V}}$ is the lower bound of the mass in the VT, which is applied to guarantee that the approximated GC $g(\mathbf{x} \vert \hat{\boldsymbol{\mu}}_i, \hat{\boldsymbol{\Sigma}}_i)$ can dominate $\mathcal{V}_i$. Here, the terms,  $\hat{\boldsymbol{\Sigma}}_i$ and its initial guess $\hat{\boldsymbol{\Sigma}}_i^0$ are applied to indicate that it is just a sub-optimal solution for the GCVT optimization problem.   


To guarantee that the variable $\boldsymbol{\Sigma}_i$ remains positive definite throughout the optimization process, we employ two practical techniques. 
\textbf{GCVT-I:} Utilizing the Cholesky decomposition, $\boldsymbol{\Sigma}_i^{-1} = \mathbf{L}_i^T \mathbf{L}_i$, we reformulate the optimization problem with respect to the matrix $\mathbf{L}_i$ instead of $\boldsymbol{\Sigma}_i$, ensuring positive definiteness by construction. 
\textbf{GCVT-II:} We express $\boldsymbol{\Sigma}_i(\alpha) = \alpha \hat{\boldsymbol{\Sigma}}_i^0$, where $0 < \alpha \leq 1$, converting the matrix-argued constrained optimization problem into a simpler scalar optimization problem with respect to the scaling factor $\alpha$. This approach further simplifies the optimization while maintaining the positive definiteness of $\boldsymbol{\Sigma}_i$.

\begin{figure}[htbp]
	\centering
	\includegraphics[width=0.485\textwidth]{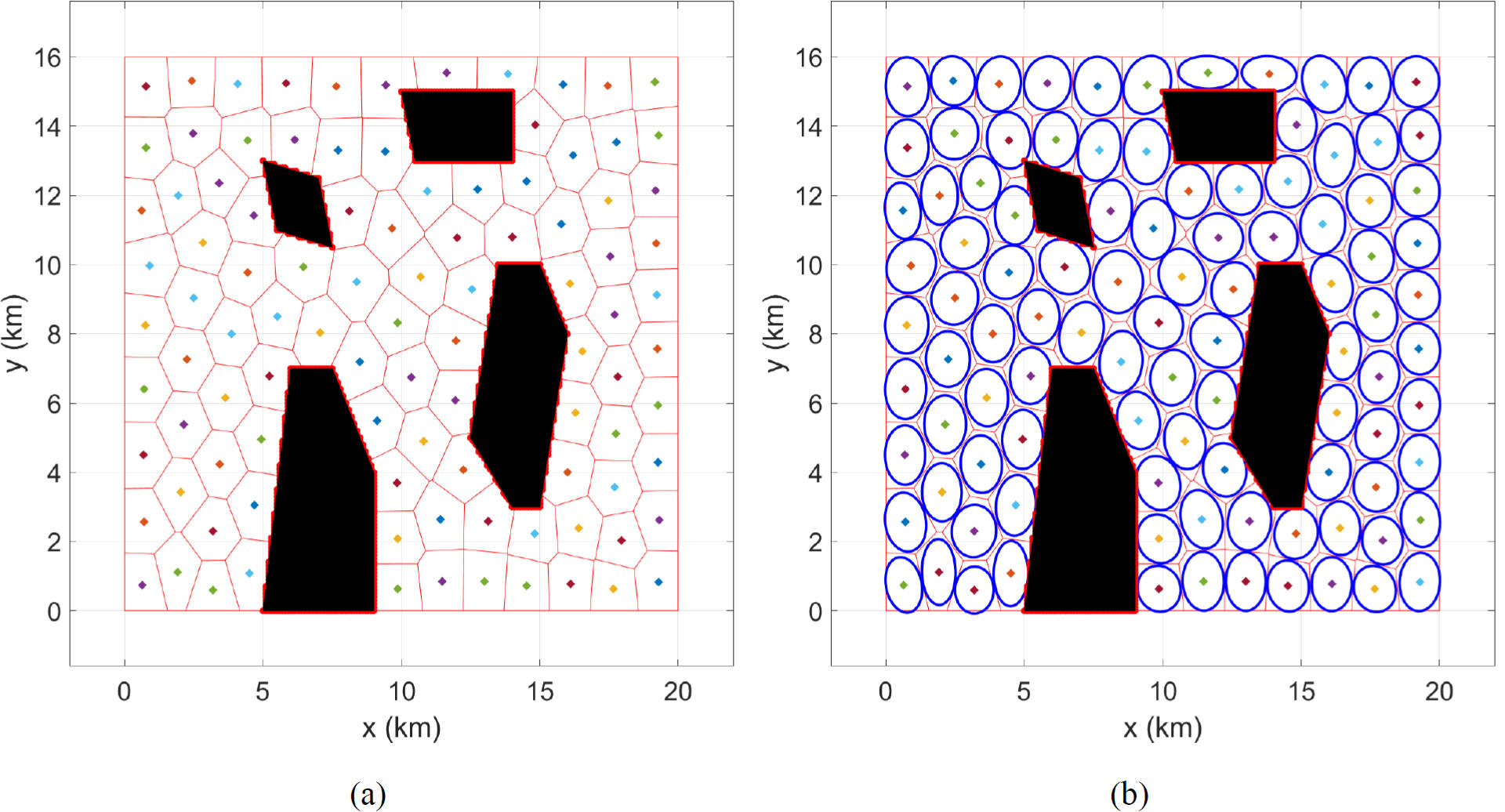} 
	\vspace{-15pt}
	\caption{The $K=100$ CVT regions are generated with traditional CVT in (a), and the heuristic Gaussian-based CVT result is shown in (b). The blue circles in (b) represent the 95\% Gaussian distribution confidence level, and the black polygons are the obstacles in the region.}
	\label{fig:cvt_result_comp}
\end{figure}

These GCVT algorithms are summarized in Algorithm-\ref{alg:cvt_gen}.  In addition, a demo of the proposed GCVT approaches are also presented in Fig. \ref{fig:cvt_result_comp} (b), where $K=100$ GCs are generated in the obstacle deployed workspace.
\begin{algorithm}[htb]
\caption{Heuristic Gaussian-based CVT-I and -II}
\label{alg:cvt_gen}
\textbf{Initialization:} \\
Number of Gaussian components $K$\\
Optimization parameters: $\eta_{\mathcal{B}}$, $\rho_{max}$ and $\kappa$, or $\alpha$ and $\eta_{\mathcal{V}}$
\textbf{Procedure: GCVT}($\mathcal{W}$, $\mathcal{X}$, $\mathcal{B}$, $K$)
\begin{algorithmic}[1]
    \State Approximate $\{( \boldsymbol{\mu}_i, \mathcal{V}_i ) \}_{i=1}^K$ using the k-means++ algorithm \cite{arthur2006k}   
    \For{ $ i=1:K $}
    	\State Approximate $\boldsymbol{\Sigma}_i$ within the VT $\mathcal{V}_i$ by solving the optimization problems formulated in (\ref{eq:argmin_GCVT_det_Sigma}) - (\ref{eq: constraint_GCVT_max_mass}) using method  
    	\textbf{GCVT-I} or \textbf{GCVT-II}.
    \EndFor
    \State \textbf{return} The approximated $\{( \boldsymbol{\mu}_i, \boldsymbol{\Sigma}_i, \mathcal{V}_i ) \}_{i=1}^K$ 
\end{algorithmic}
\end{algorithm}

\section{Numerical Implementation of CVT-Based Path Planning for VLSR Systems}
This section describes the numerical implementation of the path planning for the VLSR systems using the collocation GCs generated by the proposed GCVT algorithms. The microscopic controls for individual robots are omitted since this is not the contribution of this paper. The interested readers are referred to \cite{zhu2021adaptive, hu2024swarmprm} for implementation details. 
\subsection{Construction of Gaussian Distribution-based Graph}
Give the set of collocation GCs, $\mathbf{G}_{CVT} = \{ g^i \}_{i=1}^{K}$, where $g^i$ indicates the GC $g(\mathbf{x} \vert \boldsymbol{\mu}_i, \boldsymbol{\Sigma}_i)$ for short, and the CG sets, $\mathbf{G}_0 = \{g_0^{\imath} \}_{\imath = 1}^{N_0}$ and $\mathbf{G}_f = \{g_f^{\jmath} \}_{\jmath=1}^{N_f}$, we can construct the union set 
$\mathbf{G}$, such that 
\begin{equation}
	\mathbf{G} \triangleq \mathbf{G}_{CVT} \cup \mathbf{G}_0 \cup \mathbf{G}_f \label{eq:node_graph}
\end{equation}
We also define a set of edges between two GCs, $\mathcal{E}$, by
\begin{equation}
	\mathcal{E} \triangleq \{ (g^{\imath}, g^{\jmath})\vert g^{\imath}, g^{\jmath} \in \mathbf{G} \text{ and } W_2(g^{\imath}, g^{\jmath}) \leq d_{th}  \} \label{eq:edge_graph}
\end{equation}   
where the parameter is a user-defined distance threshold. Then, we can construct a graph of GCs, $(\mathbf{G}, \mathcal{E})$ associated with the corresponding edge cost $c_{\imath, \jmath} \triangleq W_2(g^{\jmath},g^{\imath})$. 

\subsection{Path Planning in Gaussian Distribution-based Graph}
Similar to \cite{zhu2021adaptive}, if we set the GC velocity is constant, $\nu \ll d_{th}$, then we have $\mathcal{L} (g^{\imath}, g^{\jmath}) \propto \frac{W_2(g^{\imath}, g^{\jmath})}{\nu } (\nu)^2 \propto W_2(g^{\imath}, g^{\jmath}) $. We can insert $(\big\lceil \frac{W_2(g^{\imath}, g^{\jmath})}{\nu } \big\rceil - 1)$ GCs between $g^{\imath}$ and $g^{\jmath}$ using the displacement interpolation in (\ref{eq:mu_interpolation}) and (\ref{eq:Sigma_interpolation}).   
Therefore, we can approximate a shortest path from $g_0^{\imath}$ to $g_f^{\jmath}$ based on the GC graph $(\mathbf{G}, \mathcal{E})$, denoted by $\hat{\mathcal{T}}_{0:T}^{\imath, \jmath}$. The corresponding $\mathcal{L}_{0:T}^{*\imath, \jmath}$ can also be approximated using the edge costs along the trajectory $\hat{\mathcal{T}}_{0:T}^{\imath, \jmath}$ and denoted by $\hat{\mathcal{L}}_{0:T}^{\imath, \jmath}$. Finally, we can approximate the weights $\omega_{\imath, \jmath}$ in (\ref{eq:upper_bound_objective_function}) by minimizing the following objective function,
\begin{equation}
	\{\hat{\omega}_{\imath, \jmath} \}_{\imath = 1, \jmath}^{N_0, N_T} = \underset{\{\omega_{\imath, \jmath} \}_{\imath = 1, \jmath}^{N_0, N_T}}{\argmin} \sum_{\imath = 1}^{N_0}\sum_{\jmath = 1}^{N_T} \omega_{\imath, \jmath} \hat{\mathcal{L}}_{0:T}^{\imath, \jmath} 
	\label{eq:optimization_weights}
\end{equation}
and the GMM $\wp_k$ can also be obtained by replacing $\omega_{\imath, \jmath}$ in (\ref{eq:wp_k}) by $\hat{\omega}_{\imath, \jmath}$ obtained above. 

\begin{figure}[htbp]
    \centering
    \includegraphics[width=0.5\textwidth]{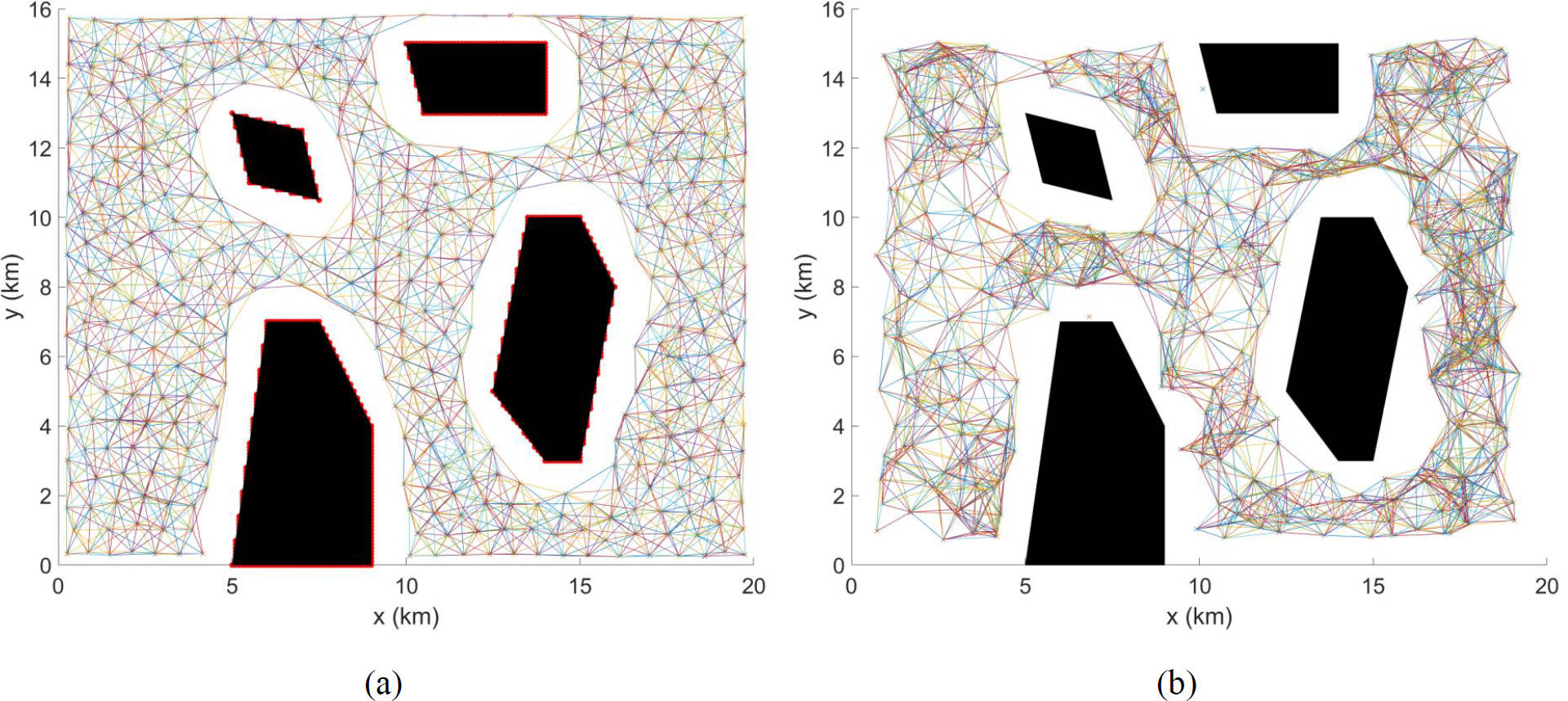}  
    \vspace{-15pt}
    \caption{An example graph visual representation generated by swarmCVT (a) and swarmPRM (b) in a region with 500 GCs, where the node is represented by the mean of $\mathbf{G}$, and the line connecting them are $\mathcal{E}$}
    \label{fig:cvt_vs_prm}
\end{figure}

The proposed path planning for the robotic swarm based on GCVT method is referred to as SwarmCVT, and the approach is summarized in Algorithm-\ref{alg:SwarmCVT}. 
Fig. \ref{fig:cvt_vs_prm} plots a graph visual representation generated by swarmCVT (a) and swarmPRM (b) by connecting the mean of the graph node given by (\ref{eq:node_graph}) and edge given by (\ref{eq:edge_graph}) when GC = $500$.

\begin{algorithm}[htb]
	\caption{SwarmCVT}
	\label{alg:SwarmCVT}
	\textbf{Initialization:} \\
	Algorithm parameters: $K$, $d_{th}$, $\nu$\\
	\textbf{Procedure: SwarmCVT}($\wp_0$, $\wp_f$, $\mathcal{W}$, $\mathcal{B}$, $\mathcal{X}$)
	\begin{algorithmic}[1]
		\State Generate the set of $K$ GCs, $\mathbf{G}_{CVT}$, and the VTs $\{\mathcal{V}_i\}_{i=1}^K$
		$\{( \boldsymbol{\mu}_i, \boldsymbol{\Sigma}_i, \mathcal{V}_i ) \}_{i=1}^K=$ GCVT($\mathcal{W}$, $\mathcal{X}$, $\mathcal{B}$, $K$)  
		\State Construct the graph, $(\mathbf{G}, \mathcal{E})$, according to (\ref{eq:node_graph}) and (\ref{eq:edge_graph}) 
		\For{ $\imath = 1:N_0$ and $\jmath = 1:N_f$}
			\State Determine the shortest GC path $\hat{\mathcal{T}}_{0:T}^{\imath, \jmath}$ in  $(\mathbf{G}, \mathcal{E})$
			\State Calculate the cost of GC path $\hat{\mathcal{L}}_{0:T}^{\imath, \jmath}$ using edge costs
		\EndFor
		\State Obtain the weights $\{\hat{\omega}_{\imath, \jmath} \}_{\imath = 1, \jmath}^{N_0, N_T}$ by solving (\ref{eq:optimization_weights})
		\State \textbf{return} The sub-optimal trajectory of GMMs, $\hat{\wp}_k$ for $k=1,\ldots, T$ using  $\{\hat{\omega}_{\imath, \jmath} \}_{\imath = 1, \jmath}^{N_0, N_T}$ and $\{\hat{\mathcal{T}}_{0:T}^{\imath, \jmath} \}_{\imath = 1, \jmath}^{N_0, N_T}$
	\end{algorithmic}
\end{algorithm}

\section{Simulation and Results}
\label{sec:Simulation_Results}
The effectiveness of the swarmCVT method is demonstrated in the following section. Due to their similarity, the ADOC and swarmPRM methods are chosen to be compared against the swarmCVT-I and swarmCVT-II methods. The swarmPRM method is tested extensively with swarmCVT methods focusing on the impact of the number of Gaussian components. 
\begin{figure}[htbp]
    \centering
    \includegraphics[width=0.485\textwidth]{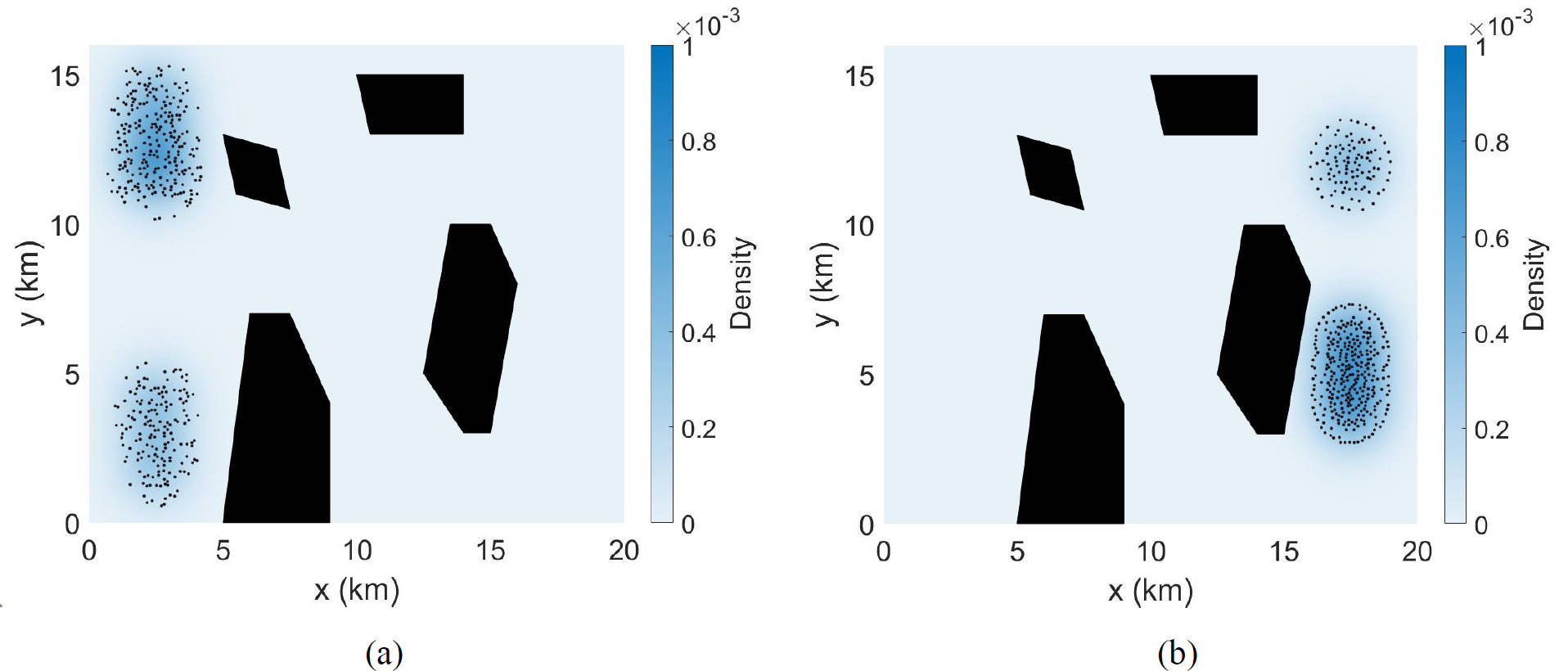}  
    \vspace{-15pt}
    \caption{VLSR system must travel from the initial distribution in (a) to the target distribution in (b), avoiding obstacles shown as polygons filled in black. Both the X and Y axes are measured in (km).}
    \label{fig:swarmCVT_progress}
\end{figure}
\subsection{Simulation Setup}
The performance of the swarmCVT approach is demonstrated on a VLSR system comprised of $N=400$ robots in the ROI defined by $\mathcal{W} = [0, W]\times [0, H]$, where $W=20$ km and $H=16$ km. Initially, all $400$ robots form an initial GMM as shown in Fig. \ref{fig:swarmCVT_progress} (a). The goal is to reach the target GMM as shown in Fig. \ref{fig:swarmCVT_progress} (b). 



The user-defined robot velocity $\nu$ is set to $5$km/h. The graph edge threshold distance is set to $d_{th} = 3$km. The obstacle overlapping parameter is set to $\eta_{\mathcal{B}} = 0.05$ to eliminate any Gaussian component overlapped by $5\%$ with the obstacles. The lower bound of VT mass is set to $\eta_{\mathcal{V}}=0.3$ to ensure at least $30\%$ of mass is within the region $\mathcal{V}$. The maximum mass is set to$\rho_{max}=0.7$ to force the highest mass in $\mathcal{V}$ is lower than $0.7$. The scale factor is set to be$\kappa=10$.

For both swarmCVT and swarmPRM methods, extensive tests are simulated on the number of Gaussian components generated in the ROI with $K=250,300,350,400,450,500$. The results are discussed in the next section, which examines the influence of this factor. 

All simulations are conducted with MATLAB code on a laptop setup with Core i7-13620H CPU @ 4.9 GHz, 64 GB RAM, and RTX4070 @ 16 RAM GPU for parallel computing. 
\begin{table}[]
\caption{TABLE I: PERFORMANCE COMPARISON }
\label{tab:result_compare}
\begin{tabular}{|c|c|c|c|c|}
\hline
\textbf{} & $T$ (min) & $D$ km)  & $W_2$-$D$ km) & Energy (J/kg)                        \\ \hline
\textbf{CVT-I}     & 9.61$\pm$0.98   & \textbf{20.11$\pm$0.20} & \textbf{20.9$\pm$0.27}   & \textbf{1.03$\pm$0.01}            \\ \hline
\textbf{CVT-II}    & \textbf{8.55$\pm$0.98}    & 20.25$\pm$0.20  & 21.17$\pm$0.32   & 1.05$\pm$0.01           \\ \hline
\textbf{PRM}  & 13.81$\pm$2.12  & 21.16$\pm$0.32 & 27.2$\pm$2.45    & 1.073$\pm$0.02          \\ \hline
\textbf{ADOC}      & 43.01$\pm$4.97  & 24.93$\pm$0.07 & 27.9859$\pm$0    & 1.286$\pm$0.01 \\ \hline
\end{tabular}
\end{table}
\subsection{Performance Comparison}
All simulations are repeated ten times for statistical reliability.  Table \ref{tab:result_compare} shows the result of all four methods on their performance on time consumption ($T$), robot average distance traveled ($D$), $W_2$ distance traveled ($W_2$-$D$), and energy cost. These results are all taken from when GC = $500$. 

\begin{figure}[htbp]
    \centering
    \includegraphics[width=0.5\textwidth]{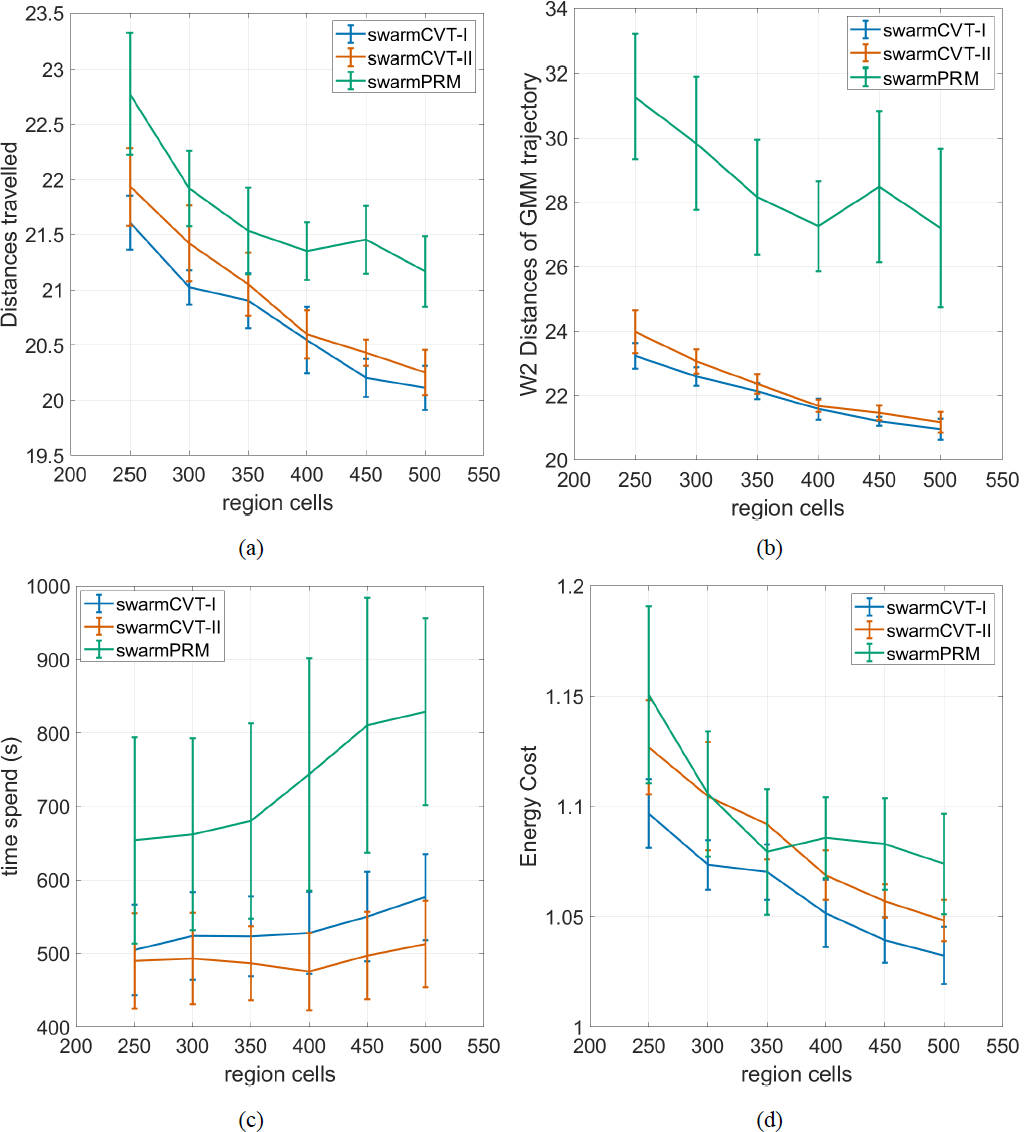}  
    \vspace{-15pt}
    \caption{Extensive comparison between swarmCVT and swarmPRM on (a) all robot average distance traveled, (b) PDF $W_2$ distance displacement, (c) time spent, and (d) energy cost under the different number of Gaussian components generated in the ROI.  Each metric is simulated ten times for each value of $K$ Gaussian component.}
    \label{fig:measurement_comb}
\end{figure}

For both swarmCVT and swarmPRM, this process is repeated ten times for each value of $K$ Gaussian components. Fig. \ref{fig:measurement_comb} (a)-(d) shows the error bars mark one standard deviation above and below the mean of each measurement. 
\begin{figure}[htbp]
    \centering
    \includegraphics[width=0.5\textwidth]{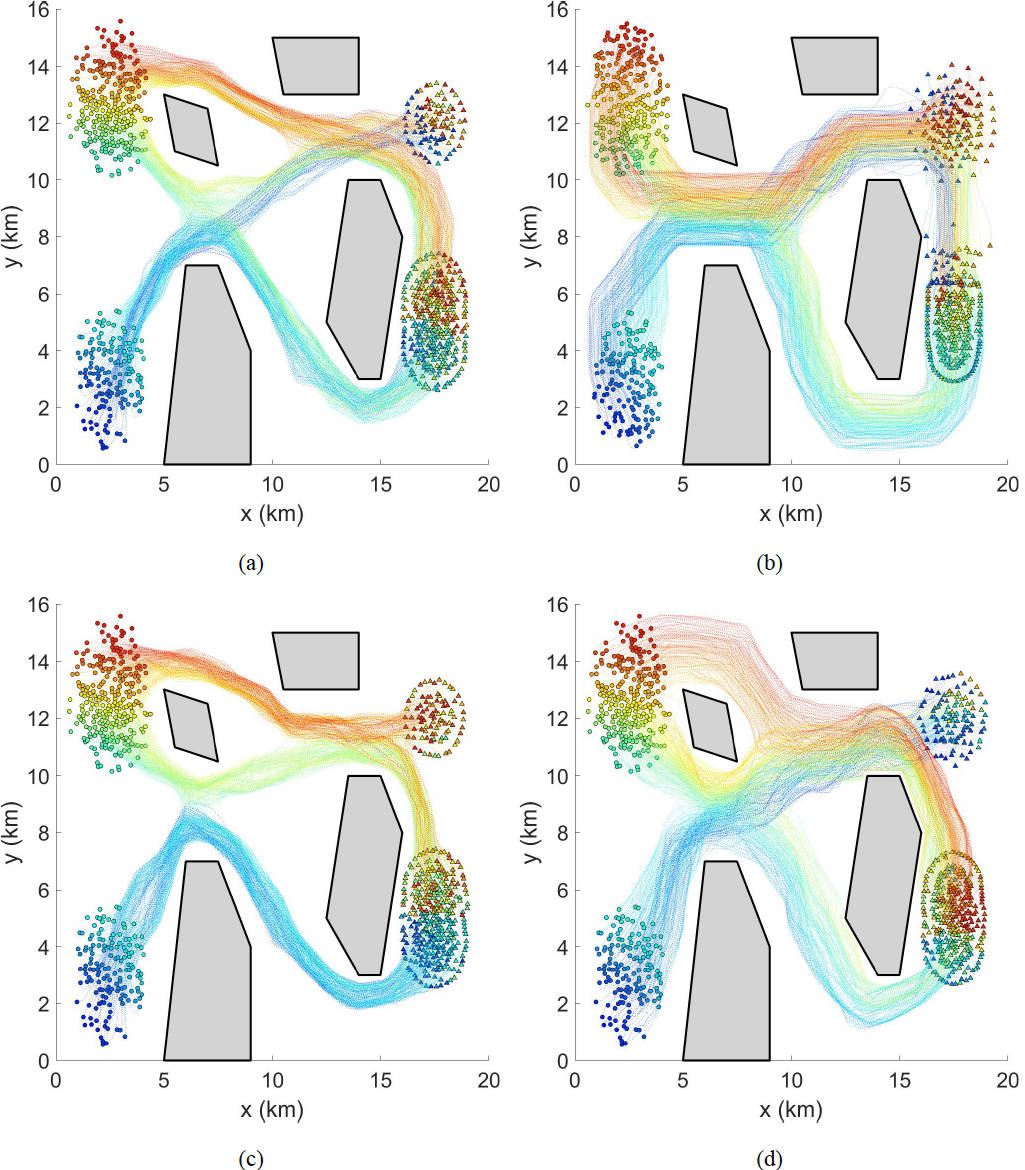}  
    \vspace{-15pt}
    \caption{VLSR trajectories from the initial (circles) to target (triangles) obtained by (a) swarmCVT-I, (b) ADOC, (c) swarmCVT-II, and (d) swarmPRM methods.}
    \label{fig:path_comb}
\end{figure}
Fig. \ref{fig:path_comb}(a)-(d) shows the VLSR trajectories obtained from the 400 robots using swarmCVT-I, swarmCVT-II, swarmPRM, and ADOC. All methods can reach the target, but it can be examined that swarmCVT methods outperform swarmPRM and ADOC on all four measurements. 


\section{Conclusion and Future Works}
In this paper, we propose a novel Gaussian distribution-based centroidal Voronoi tessellation (GCVT) concept inspired by traditional CVT and develop two approximation methods to solve the GCVT problems. Building on this foundation, we developed SwarmCVT, a new variant of the adaptive distribution optimal control (ADOC) algorithm for path planning in environments with known obstacle layouts. We evaluated the performance of SwarmCVT against ADOC and SwarmPRM, another variant of ADOC. Simulation results demonstrated that SwarmCVT significantly outperforms both ADOC and SwarmPRM. Our work on GCVT opens up new research directions, and future work will explore additional applications and further improvements to the GCVT framework.

\addtolength{\textheight}{-12cm}   









\bibliographystyle{IEEEtran}
\bibliography{ref,ref_Ping}

\end{document}